\documentclass[letterpaper, 10 pt, journal, twoside]{IEEEtran}
\ifCLASSINFOpdf
\else
\fi

\usepackage{amsthm}
\usepackage{amsmath,amssymb,amsfonts}
\usepackage{booktabs}
\usepackage{float}
\usepackage{algorithm}
\usepackage{multicol}
\usepackage{duckuments}
\usepackage{graphicx}
\usepackage{xr}

\usepackage{enumitem}
\usepackage{etoolbox}
\usepackage{mathrsfs}
\usepackage{multirow}
\usepackage{flushend}
\usepackage{bbding}
\usepackage{color}
\usepackage[table,xcdraw,dvipsnames]{xcolor}
\usepackage{threeparttable}
\usepackage{lipsum}
\usepackage{algorithmic}
\usepackage{url}

\definecolor{bshade}{rgb}{0.55,0.75,0.95}
\title{\LARGE \bf
Breaking the Static Assumption: A Dynamic-Aware LIO Framework Via Spatio-Temporal Normal Analysis
}

\begin{document}
%
\title{Breaking the Static Assumption: A Dynamic-Aware LIO Framework Via Spatio-Temporal Normal Analysis}
%
%
%


\author{Zhiqiang Chen$^{1}$, Cedric Le Gentil$^{2}$, Fuling Lin$^{1}$, Minghao Lu$^{1}$, Qiyuan Qiao$^{1}$, \\ Bowen Xu$^{1}$, Yuhua Qi$^{3}$, and Peng Lu$^{1}$%
\thanks{Manuscript received: Apri 17, 2025; Revised July 22, 2025, Year; Accepted October 12, 2025. This paper was recommended for publication by Editor Sven Behnke upon evaluation of the Associate Editor and Reviewers' comments.
(Corresponding author: Peng Lu)}
\thanks{$^{1}$Zhiqiang Chen, Fuling Lin, Minghao Lu, Qiyuan Qiao, Bowen Xu and Peng Lu are with Department of Mechanical Engineering, The University of Hong Kong, Hong Kong SAR, China.
        {\tt\footnotesize lupeng@hku.hk}}%
\thanks{$^{2} $Cedric Le Gentil is with Autonomous Space Robotics Lab, University of Toronto Institute for Aerospace Studies (UTIAS), Ontario, Canada.}
\thanks{$^{3} $Yuhua Qi is with The School of Systems Science and Engineering, Sun Yatsen University, Guangzhou, China}%
\thanks{Digital Object Identifier (DOI): see top of this page.}
}
%
%

\markboth{IEEE Robotics and Automation Letters. Preprint Version. Accepted OCTOBER, 2025}
{CHEN \MakeLowercase{\textit{et al.}}: Breaking the Static Assumption: A Dynamic-Aware LIO Framework Via Spatio-Temporal Normal Analysis} 

%



\maketitle

\begin{abstract}
This paper addresses the challenge of Lidar-Inertial Odometry (LIO) in dynamic environments, where conventional methods often fail due to their static-world assumptions. Traditional LIO algorithms perform poorly when dynamic objects dominate the scenes, particularly in geometrically sparse environments. Current approaches to dynamic LIO face a fundamental challenge: accurate localization requires a reliable identification of static features, yet distinguishing dynamic objects necessitates precise pose estimation. Our solution breaks this circular dependency by integrating dynamic awareness directly into the point cloud registration process. We introduce a novel dynamic-aware iterative closest point algorithm that leverages spatio-temporal normal analysis, complemented by an efficient spatial consistency verification method to enhance static map construction. Experimental evaluations demonstrate significant performance improvements over state-of-the-art LIO systems in challenging dynamic environments with limited geometric structure. 
The code and dataset are available at \url{https://github.com/thisparticle/btsa}.
\end{abstract}


%
\IEEEpeerreviewmaketitle

\section{INTRODUCTION}
\begin{figure}[t]
  \centering
  \includegraphics[width=0.48\textwidth]{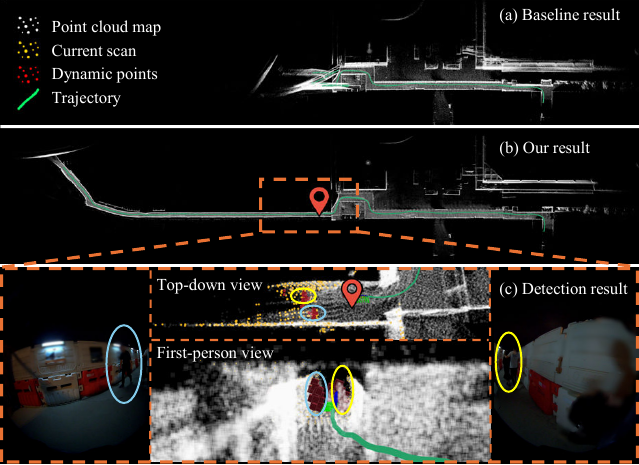}
  \caption{Comparison of our method against FAST-LIO2 \cite{xu2022fast} for mapping and trajectory estimation. Colored ellipses highlight detected dynamic objects. Our approach reduces localization drift through effective dynamic object handling, especially in challenging geometric environments.}
  \label{fig:teaser}
  \vspace{-0.6cm} 
\end{figure}

Lidar-Inertial Odometry (LIO) has become essential for precise localization and mapping across robotics applications by integrating detailed 3D point clouds with high-frequency motion data. Despite its effectiveness, conventional LIO systems \cite{shan2020lio, legentil2021in2laama, xu2022fast, chen2024ig} operate under a static environment assumption that rarely holds in practice. When moving objects like pedestrians and vehicles appear in the sensor field of view, they introduce significant errors in both pose estimation and map construction. These errors stem from registration algorithms such as Iterative Closest Point (ICP) \cite{pomerleau2015review} mistakenly using dynamic elements as fixed reference points. This problem becomes particularly acute in environments where dynamic objects predominate or where static geometric features are limited, leading to substantial localization failures as illustrated in Fig. \ref{fig:teaser}.

Existing approaches to this challenge primarily adopt a pre-processing strategy—removing dynamic objects before registration—rather than addressing the fundamental registration algorithm limitations. Geometric filtering methods \cite{wu2024observation, yuan2024lidar} rely on structural assumptions that restrict their applicability across diverse environments. Meanwhile, learning-based techniques \cite{jia2024trlo} only detect predefined object categories rather than all dynamic points, while requiring substantial training data and showing limited generalization capabilities. These approaches create a circular dependency: accurate state estimation requires reliable static feature identification, yet effective dynamic object detection depends on precise pose information.
Moreover, evaluation protocols for dynamic LIO systems have largely relied on autonomous driving datasets from geometrically-rich urban environments \cite{hsu2021urbannav, chen2023ecmd}. The abundance of static features in these settings often minimizes dynamic object impact, understating the true challenges of localization in predominantly dynamic environments. This evaluation bias has led to an underappreciation of the dynamic localization problem's significance in practical applications.

In this paper, we address these critical limitations by proposing a novel framework that explicitly models and accounts for dynamic elements in the environment during point cloud registration. Our approach breaks the circular dependency between state estimation and dynamic object detection by integrating spatio-temporal normal analysis directly into the registration process, maintaining accurate trajectory estimation even in highly dynamic scenes with limited geometric structure. Our contributions include:

\begin{itemize}
    \item A novel dynamic-aware ICP algorithm for robust point cloud registration in dynamic environments that addresses the circular dependency between localization and dynamic object detection
    \item The empirical demonstration of our method's effectiveness in truly challenging dynamic environments, including geometrically degraded scenes and scenarios dominated by dynamic objects.
    \item A computationally efficient spatial consistency verification approach that enhances static map construction.
    \item The open-source release of our code and a new dataset featuring challenging dynamic environments that current methods struggle to handle effectively.
\end{itemize}

\section{RELATED WORKS}
In the literature, the processes of ego-motion estimation and dynamic object detection are generally performed one after the other, and sometimes iteratively.
In this section, we provide a brief overview of the different methods for Lidar-inertial state estimation and dynamic object detection. 

\subsection{Lidar-inertial State Estimation}
Lidar-inertial odometry combines lidar's precise geometric measurements with IMU's high-frequency motion data to enable robust state estimation. Early point-based approaches like LOAM \cite{zhang2014loam} and its variants \cite{shan2018lego, lin2020loam} extract geometric features for real-time performance. Subsequent developments have improved computational efficiency through compact environmental representations including voxels \cite{yuan2022efficient} and surfels \cite{ramezani2022wildcat}. Modern registration techniques have advanced beyond simple distance metrics to incorporate probabilistic \cite{chen2024ig}, semantic \cite{chen2019suma++}, and intensity-based \cite{pfreundschuh2024coin} approaches for enhanced matching accuracy.

Despite these advances, conventional LIO systems struggle when operating in dynamic environments. While outlier rejection mechanisms and robust loss functions partially mitigate the impact of moving objects in moderately dynamic scenes, they become insufficient when a significant portion of the sensor data corresponds to dynamic elements.
Recent efforts have attempted to address this limitation by integrating learning-based detection modules with odometry systems \cite{jia2024trlo}. Although these approaches improve performance in dynamic scenes, they suffer from limited generalizability in highly dynamic environments. Alternative geometric detection methods \cite{yuan2024lidar, Lichtenfeld2024ddlo} show promise but remain constrained by specific assumptions about object geometry and require dense point cloud data, limiting their practical applicability.
A critical limitation of existing approaches is their treatment of dynamic object detection as a separate processing step, creating a circular dependency between accurate pose estimation and reliable dynamic object identification. Some recent works \cite{zhu2024limot, lin2023asynchronous} attempt simultaneous odometry and dynamic object tracking by treating moving objects as landmarks. While these methods demonstrate improved accuracy in structured driving scenarios, they require smooth object trajectories and rely on deep learning detectors which limit their applicability to new environments.

\begin{figure}[t]
  \centering
    \includegraphics[width=0.37\textwidth]{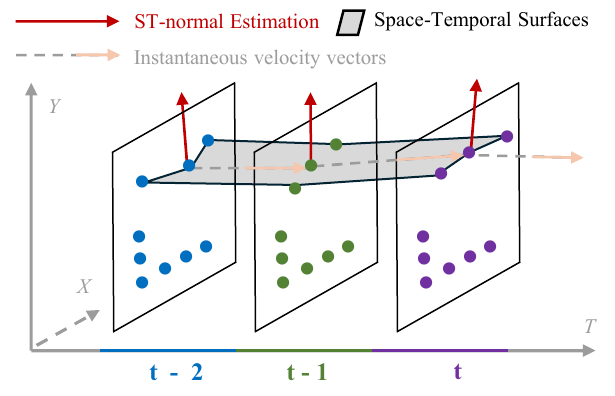}
  \caption{Space-Time Velocity Vectors in a 2D Example. The instantaneous velocity vectors (in pink) are tangent to the space-time surfaces generated by motion, while the space-time normals (in red) are perpendicular to these surfaces. Each instantaneous velocity vector can be decomposed into a temporal component, determined by the frame rate, and a spatial component representing the underlying motion of points.}
  \label{fig:st_normal}
\end{figure}

\subsection{Dynamic object detection}
Dynamic object detection methodologies can be categorized into several distinct approaches. Offline removal methods \cite{Lim2021ERASORER, Jia2024BeautyMapBA, lim2023erasor2} achieve impressive results by leveraging information from the complete accumulated map, but inherently sacrifice real-time operation capability.

Online approaches are primarily divided into learning-based and geometric-based paradigms. Learning-based methods employ neural network architectures to segment dynamic elements \cite{wang2023insmos, mersch2023building}, typically processing point cloud frames enhanced with temporal information. Despite their effectiveness, these approaches require extensive training datasets and often exhibit diminished performance when encountering novel object categories or sensor configurations.

Geometric-based techniques operate without labeled training data and include two primary subtypes. FreeSpace-based methods \cite{schmid2023dynablox} maintain environmental representations through updated voxel occupancy states, while difference-based approaches \cite{wu2024observation, wu2024moving} identify dynamic elements by analyzing occlusion patterns between sequential scans. Although these methods avoid the data dependencies of learning-based approaches, they typically struggle with complex occlusion scenarios and angular incidence ambiguities. Recent spatio-temporal normal frameworks \cite{falque2023dynamic,legentil2024realtime} encode dynamic characteristics through surface geometry and motion relationships, but perform detection as post-processing after pose estimation, requiring future frames and focusing on static mapping rather than localization.

A fundamental challenge in this domain is that most existing methods presuppose accurate sensor pose information, leading to the aforementioned estimation/detection inter-dependence. Our work addresses this by integrating spatio-temporal normal analysis directly into the iterative registration loop, enabling simultaneous real-time refinement of pose estimates and dynamic point classification for the current frame.

\begin{figure*}[t]
  \centering
    \includegraphics[width=0.95\textwidth]{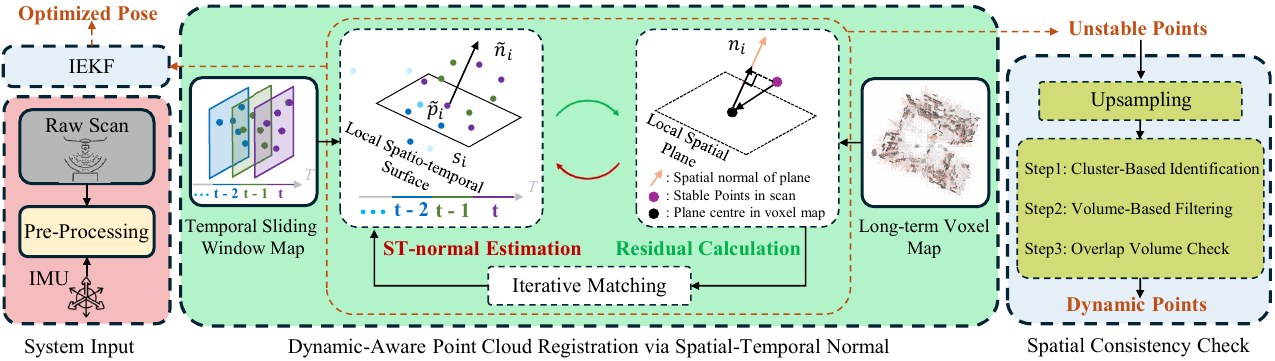}
    \vspace{-0.25cm} 
  \caption{An Overview of the Proposed Framework for Dynamic-Aware Point Cloud Registration and Static Map Construction. The Pre-Processing step undistorts the point cloud, which is then input into the dynamic ICP loop with a Temporal Sliding Window Map. The optimized pose, eliminating dynamic elements, is derived using the proposed spatial-temporal estimation and selective update procedure. The static map is constructed by filtering out false positives from unstable points through a spatial consistency check.
  }
  \label{fig:Our_Pipeline}
  \vspace{-0.5cm} 
\end{figure*}

\section{PRELIMINARIES}

As this work leverages the concept of spatio-temporal normals for dynamic point detection, we derive here a rigorous definition of the spatio-temporal normals and their link to velocity beyond the original work in~\cite{falque2023dynamic}. Let us denote a point cloud captured at time $t^j$ as $\mathcal{P}^j = \{\mathbf{p}_i^j\}$, where each point $\mathbf{p}_i^j$ is expressed in the fixed reference frame $\mathcal{F}_W$. To incorporate temporal information, we augment each point with its acquisition time, creating a space-time representation $\tilde{\mathcal{P}}^j = \{(\mathbf{p}_i^j, t^j) | \mathbf{p}_i^j \in \mathbb{R}^3, t^j \in \mathbb{R}\}$. Throughout this paper, we use tilde notation $(\tilde{})$ to distinguish space-time elements from purely spatial ones.

\subsubsection{Spatio-Temporal Surface and Its Normals}
The spatio-temporal surface $\mathcal{S}$ is represented by the implicit function $g(x, y, z, t) = 0$ in 4D space. At any point $\tilde{\mathbf{p}}_i^j = (x_i^j, y_i^j, z_i^j, t^j)^\top$ on this surface, the normal vector equals the gradient $\nabla g = (a, b, c, d)^\top$, where components correspond to partial derivatives with respect to spatial coordinates and time.
For a point moving along this surface, differentiating $g(x, y, z, t) = 0$ with respect to time yields $ a v_x + b v_y + c v_z + d = 0$, 
where $(v_x, v_y, v_z)$ represents the point's velocity. Rearranging this equation, we obtain:
\begin{equation}
d = -(a v_x + b v_y + c v_z).
\label{eq:temporal_component}
\end{equation}

This equation demonstrates that the temporal component $d$ of the 4D normal vector quantifies the surface's temporal evolution, directly relating to the projection of instantaneous velocity onto the spatial gradient.
As shown in Fig. \ref{fig:st_normal}, the velocity vector field aligning consecutive frames should be perpendicular to the surface normal field. This property enables us to identify dynamic points by comparing the temporal component $d$ of the 4D normal with a predefined threshold.

\subsubsection{Spatio-Temporal Normal Estimation}
To estimate the four-dimensional surface normal $\tilde{\mathbf{n}} = (a, b, c, d)$ at point $\mathbf{p}_i^j$, we fit a tangent hyperplane to the local point cloud neighborhood $\mathcal{N}_i^j$. 
Concretely, $\tilde{\mathbf{n}}$ corresponds to the eigenvector associated with the smallest eigenvalue in the eigen decomposition of the covariance matrix
\begin{equation}
    \operatorname{cov}_i^j = \frac{1}{\|\mathcal{N}_i^j\|} \sum_{\mathbf{p}_u^v \in \mathcal{N}_i^j} \left( \begin{bmatrix} \mathbf{p}_u^v \\ t_u^v \end{bmatrix} - \mathbf{m}_i^j \right) \left( \begin{bmatrix} \mathbf{p}_u^v \\ t_u^v \end{bmatrix} - \mathbf{m}_i^j \right)^{\top},
\end{equation}
where $\mathbf{m}_i^j = \frac{1}{\|\mathcal{N}_i^j\|} \sum_{\mathbf{p}_u^v \in \mathcal{N}_i^j} \begin{bmatrix} {\mathbf{p}_u^v}^\top & t_u^v \end{bmatrix}^\top$ is the neighborhood's centroid.
Please note that accurate hyperplane fitting requires neighboring points to be uniformly distributed in both spatial and temporal dimensions within a sufficiently small neighborhood. While spatial uniformity is typically achievable, temporal uniformity presents challenges when processing live data, as it would require future data.
As illustrated in Fig.~\ref{fig:scc}(a), newly observed points can exhibit apparent non-zero velocities, potentially leading to false dynamic detections.
However, such points are generally not reliable for the task of scan registration.
In this work, we categorize both truly dynamic and unreliable points as ``unstable points".

\section{METHOD}

\subsection{Overview}
Let us consider an IMU and 3D LiDAR rigidly mounted.
Inspired by~\cite{xu2022fast}, the proposed method estimates the system's pose $\mathbf{T}^j$ and velocity $\mathbf{v}^j$ at time $t_j$ as well as the IMU biases using an Iterated Extended Kalman Filter (IEKF).
We denote the system's overall state with  $\mathbf{x}^j$.
As illustrated in Fig.~\ref{fig:Our_Pipeline}, the method consists of three main components: the input data pre-processing, our novel dynamic-aware point cloud registration algorithm, and the associated static map building.
Based on IMU preintegration, the pre-processing step propagates the previous state estimate $\mathbf{x}^{j-1}$ to obtain a prior for $\mathbf{x}^j$ and corrects the motion distortion of the lidar point cloud.
Before performing registration and static map building, both detailed in the next subsections, the point cloud is downsampled (voxel-based) to reduce the computational burden of the subsequent steps.

\subsection{Dynamic-Aware Point Cloud Registration}
Standard ICP registration consists of iteratively performing data association and geometric transformation estimation between two point clouds until convergence.
Generally, ICP leverages a simple distance-based outlier rejection to omit erroneous associations.
The proposed dynamic-aware point cloud registration extends the standard point-to-plane ICP to explicitly account for the presence of dynamic objects in the scene by extending the outlier rejection step based on spatio-temporal normal computation.
The approach is summarized in Algorithm \ref{alg:dicp-algorithm}.

\begin{algorithm}[t]
\footnotesize
\renewcommand{\algorithmicrequire}{\textbf{Input:}}
\renewcommand{\algorithmicensure}{\textbf{Output:}}
\caption{Dynamic-Aware Point Cloud Registration}
\label{alg:dicp-algorithm}
\begin{algorithmic}[1]
\REQUIRE Point cloud $\tilde{\mathcal{P}}$, Temporal sliding window map $\mathcal{M}_t$, Long-term voxel map $\mathcal{M}_v$, Initial pose $\hat{\mathbf{x}}^0$
\ENSURE Optimized pose $\hat{\mathbf{x}}^\star$
\STATE $iter \leftarrow 0$, $converged \leftarrow false$
\WHILE{$iter < max\_iter$ AND $\neg converged$}
    \STATE Transform $\tilde{\mathcal{P}}$ to world coordinates using $\hat{\mathbf{x}}^{iter}$
    \STATE Compute spatio-temporal normal vectors $\tilde{\mathbf{n}}$ using $\mathcal{M}_t$
    \STATE Identify stable points $\tilde{\mathcal{P}}_s$ based on $\tilde{\mathbf{n}}$
    \STATE Find corresponding surfaces in $\mathcal{M}_v$ for each point in $\tilde{\mathcal{P}}_s$
    \STATE Optimize pose $\hat{\mathbf{x}}^{iter+1}$ using point-to-plane error metric
    \STATE $converged \leftarrow \|\hat{\mathbf{x}}^{iter+1} - \hat{\mathbf{x}}^{iter}\| < \epsilon$
    \STATE $iter \leftarrow iter + 1$
\ENDWHILE
\STATE $\hat{\mathbf{x}}^\star \leftarrow \hat{\mathbf{x}}^{iter}$
\STATE Update maps $\mathcal{M}_t$ and $\mathcal{M}_l$ with $\tilde{\mathcal{P}}$ and $\hat{\mathbf{x}}^\star$
\RETURN $\hat{\mathbf{x}}^\star$
\end{algorithmic}
\end{algorithm}

\subsubsection{Dynamic-Aware ICP}
To estimate the state $\mathbf{x}^j$ from the current scan $\tilde{\mathcal{P}}^j$, we fit local spatiotemporal planes to each point in $\tilde{\mathbf{p}}^j_i$ in $\tilde{\mathcal{P}}^j$ based on a temporal sliding window map $\mathcal{M}_t$ and the initial guess of $\mathbf{x}^j$.
The local temporal map $\mathcal{M}_t$ contains recent lidar data registered in a single reference frame.
Each point in the current frame is classified as ``stable" or ``unstable" based on the temporal component $d$ of the spatio-temporal normals (cf.~\eqref{eq:temporal_component}) using a predefined threshold $d_{thr}$.
Unstable points are discarded as they correspond to a dynamic object or a newly observed area and are not reliable for pose estimation.

The stable points are then used for state estimation with respect to a global map $\mathcal{M}_v$ that only contains static elements of the environment.
The state estimation corresponds to minimizing both IMU preintegration residuals and point-to-plane distances between $\mathcal{P}^j$ and $\mathcal{M}_v$:
\begin{equation}
    \hat{\mathbf{x}}^j = \underset{\mathbf{x}}{\arg\min} \left(  \| \mathbf{r}_{\text{IMU}} \|_{\Sigma}^2 + \sum_{i} \| \mathbf{n}_i^T ({}^W \! \mathbf{p}_i-\mathbf{q}_i) \|^2_{\mathbf{R}_i} \right)
    \label{eq:optimization},
\end{equation}
where $\mathbf{r}_{\text{IMU}}$ denotes the IMU preintegration residuals with covariance $\Sigma$, $\mathbf{n}_i$ is the normal vector of the corresponding plane at target point $\mathbf{q}_i$ in $\mathcal{M}_v$, and $\mathbf{R}_j$ represents the LiDAR measurement covariance matrix.
This process is repeated until the state estimate converges.
Note that evaluating the point's dynamicity through spatio-temporal normal analysis is done at each iteration, thus jointly solving the problems of state estimation and dynamic object detection. 

\subsubsection{Dual-Map Architecture for Registration}
As mentioned in the previous paragraph, the proposed dynamic-aware registration relies on a temporal sliding window map $\mathcal{M}_t$ and a long-term voxel map $\mathcal{M}_v$.
This approach balances computational efficiency and accuracy: the local map $\mathcal{M}_t$ needs to be temporally dense for accurate spatio-temporal estimation, while the voxel-based long-term map $\mathcal{M}_v$ ensures global consistency of the pose estimate efficiently.

The temporal sliding window map ($\mathcal{M}_t$) retains approximately two seconds of recent LiDAR point cloud, providing the temporal context necessary for calculating spatio-temporal surface normal. We implement this using an iKdTree structure \cite{cai2021ikd} paired with a double-ended queue. After pose estimation, each new point cloud frame is transformed to world coordinates and simultaneously added to both the queue and the k-d tree. Correspondingly, the oldest frame is removed from the queue and its points are deleted from the tree, ensuring the map remains temporally current.
In parallel, we maintain a long-term voxel map ($\mathcal{M}_v$) that stores the spatial planes and their normal following the VoxelMap approach \cite{yuan2022efficient}, which provides a more extensive spatial reference for accurate registration. This dual-map architecture effectively addresses the competing requirements of our system: temporal proximity for spatio-temporal normal calculation and spatial richness for robust localization.

\subsubsection{Threshold Selection for Normal Vectors}
The temporal component $d$ in a spatio-temporal normal vector $\tilde{\mathbf{n}} = (a, b, c, d)$ indicates surface motion, with $d=0$ for static points. We establish a threshold $d_{thr}$ by examining the angle $\theta$ between $\tilde{\mathbf{n}}$ and its spatial projection $(a, b, c, 0)$. This angle represents the deviation into the temporal dimension caused by motion and is calculated as:
$\cos\theta = \frac{a^2 + b^2 + c^2}{a^2 + b^2 + c^2 + d^2} = 1 - \frac{d^2}{|\tilde{\mathbf{n}}|^2}$.
This approach offers a physically interpretable threshold selection method. For example, setting $\theta_{thr}=5.7^\circ$ yields $|d| \approx 0.1$, providing an intuitive parameter that directly relates to observable motion characteristics.

\subsection{Static Map Building via Spatial Consistency Check}
\begin{figure}[t]
  \centering
\includegraphics[width=0.47\textwidth]{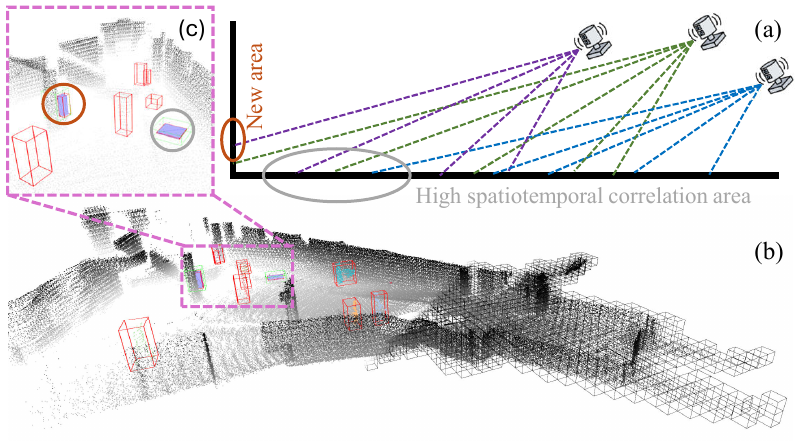}
\vspace{-0.25cm} 
  \caption{Spatial Consistency Check. Figure (a) illustrates the causes of false positives in dynamic detection by using spatiotemporal normal vectors over a specified time window. Figure (b) provides an example demonstrating our approach to eliminating false positives based on the spatial consistency principle. Figure (c) is a zoomed-in view of Figure (b) that highlights two distinct cases of false positives.}
  \label{fig:scc}
  \vspace{-0.5cm} 
\end{figure}
When analyzing spatio-temporal normal vectors from the sliding time map, we identify unstable points representing both dynamic objects and false positives. 
Fig. \ref{fig:scc}(a) illustrates how these false positives emerge from two primary sources: newly observed regions appearing for the first time in the map, and static areas where scan patterns change significantly between consecutive frames. In both cases, the calculation erroneously produces non-zero temporal components, misclassifying static points as dynamic. This misclassification stems from inherent LiDAR sensing limitations and data sparsity rather than actual movement in the environment.

To address false positives in our static mapping process, we implemented a spatial consistency detection approach based on a key observation: dynamic and static points exhibit distinct spatial distribution patterns—dynamic points cluster together while static points are typically surrounded by other static points. Our method first upsamples unstable points from the voxel-downsampled point cloud through nearest neighbor search to identify potentially unstable adjacent points. We then apply DBSCAN clustering \cite{ester1996density} to effectively eliminate isolated false positives.
Each resulting cluster is enclosed in a bounding box, as shown in Fig.~\ref{fig:scc}(b). We discard oversized clusters as outliers and introduce a lightweight sliding voxel map, $\mathcal{M}_{scc}$, that maintains a short-term record of static areas near the sensor. By analyzing the volumetric overlap between candidate dynamic clusters and $\mathcal{M}_{scc}$, we can differentiate between genuine dynamic objects (minimal overlap) and false positives such as newly observed static regions (significant overlap), as illustrated in Fig.~\ref{fig:scc}(c). This spatial consistency check substantially improves our static map's accuracy by reliably distinguishing true dynamic elements from detection artifacts.

\section{EXPERIMENTS}
We evaluated our approach in challenging dynamic environments, with particular focus on scenarios containing numerous moving objects. Our experiments assessed odometry accuracy, static map construction capabilities across multiple LiDAR sensor types, and computational efficiency. All evaluations were conducted on a computer with an Intel i5-12490 CPU, 32GB RAM, running Ubuntu 22.04.

\begin{table}
\caption{DATASETS OF ALL SEQUENCES FOR EVALUATION}

\centering{}%
\resizebox{0.47\textwidth}{!}{
\begin{tabular}{ccccc}
\toprule 
Scene Type & Abbreviation & Sequence Name & Path Length & Duration\tabularnewline
\midrule
\multirow{7}{*}{GR} & D\_1 & Promenade01 & 357m & 278s\tabularnewline
 & D\_2 & Promenade02 & 267m & 195s\tabularnewline
 & E\_1 & \cite{chen2023ecmd} - Dense\_street\_day\_difficult\_circle & 2187m & 803s\tabularnewline
 & E\_2 & \cite{chen2023ecmd} - Dense\_street\_day\_medium\_a & 356m & 120s\tabularnewline
 & E\_3 & \cite{chen2023ecmd} - Dense\_street\_day\_medium\_b & 278m & 125s\tabularnewline
 & U\_1 & \cite{hsu2021urbannav} - UrbanNav-HK-Data20190428 & 1997m & 487s\tabularnewline
 & U\_2 & \cite{hsu2021urbannav} - UrbanNav-HK-Data20200314 & 1212m & 300s\tabularnewline
\midrule
\multirow{3}{*}{GD} & G\_1 & \cite{chen2024heterogeneous} - Bridge01 & 3966m & 382s\tabularnewline
 & G\_2 & \cite{chen2024heterogeneous} - Urban\_tunnel03 & 6749m & 335s\tabularnewline
 & D\_3 & Promenade03 & 276m & 200s\tabularnewline
\midrule
\multirow{3}{*}{DOD} & D\_4 & MountainTopPark01 & 393m & 289s\tabularnewline
 & D\_5 & MountainTopPark02 & 362m & 270s\tabularnewline
 & D\_6 & MountainTopPark03 & 266m & 226s\tabularnewline
\bottomrule
\end{tabular}
}
\label{Tab:dataset}
\end{table}
\begin{figure}[t]
  \centering
\includegraphics[width=0.47\textwidth]{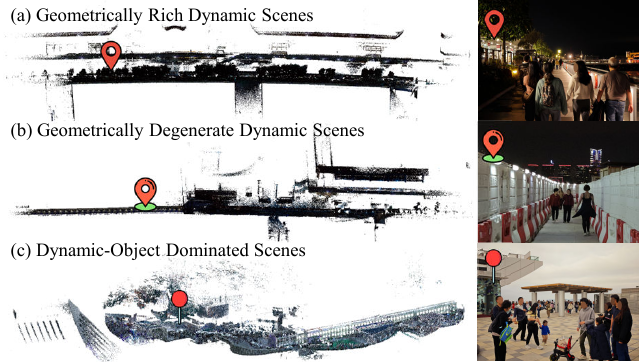}
  \caption{Examples of 3D Point Cloud and Scene Image Visualizations from Self-Collected Dataset.}
  \label{fig:dynamic_dataset_img}
  \vspace{-0.3cm} 
\end{figure}

\subsection{Odometry}

\begin{figure*}[t] \centering
\makebox[0.32\textwidth]{\footnotesize Promenade03}
    \makebox[0.32\textwidth]{\footnotesize Bridge01}
    \makebox[0.32\textwidth]{\footnotesize Urban\_tunnel03}
    \\
    \includegraphics[width=0.32\textwidth]{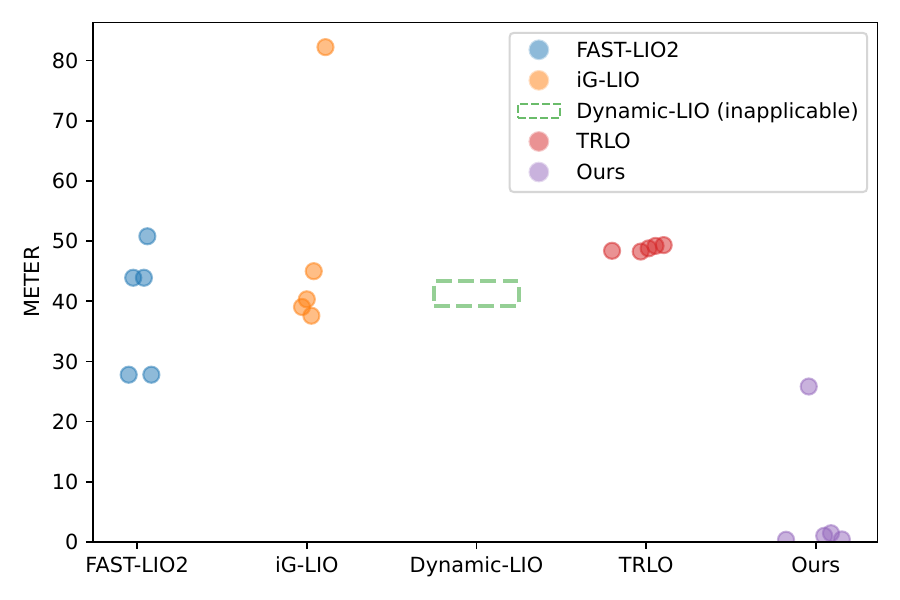}
    \includegraphics[width=0.32\textwidth]{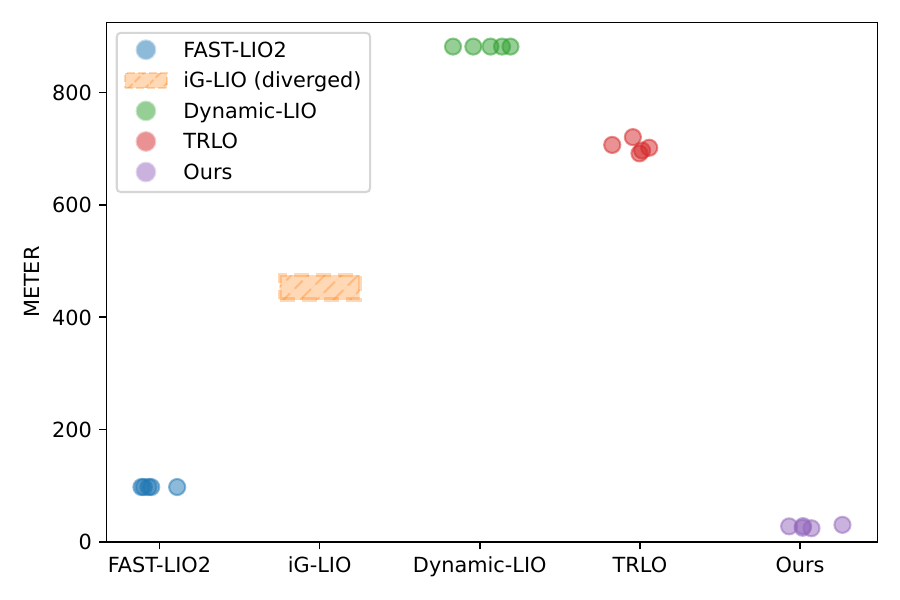}
    \includegraphics[width=0.32\textwidth]{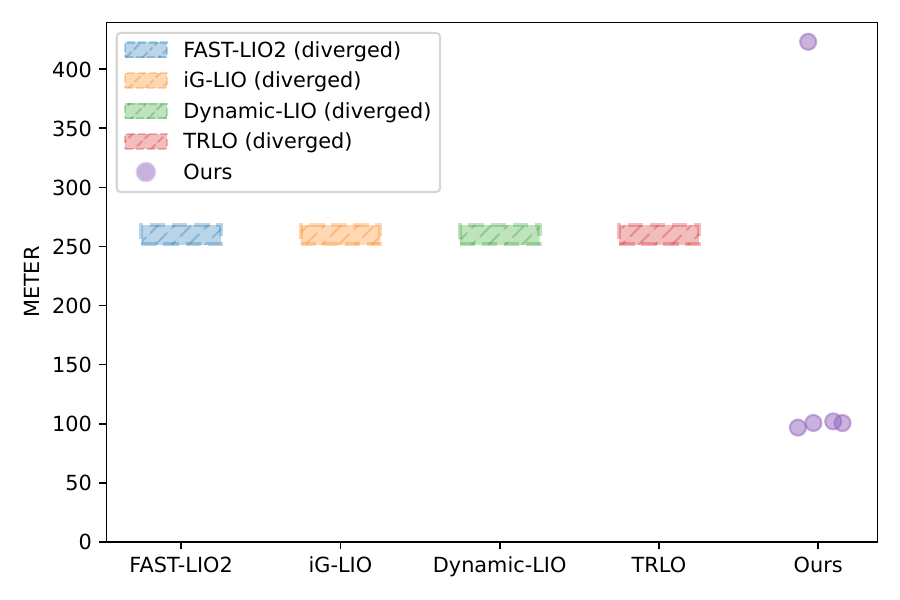}
    \\  
    \includegraphics[width=0.32\textwidth]{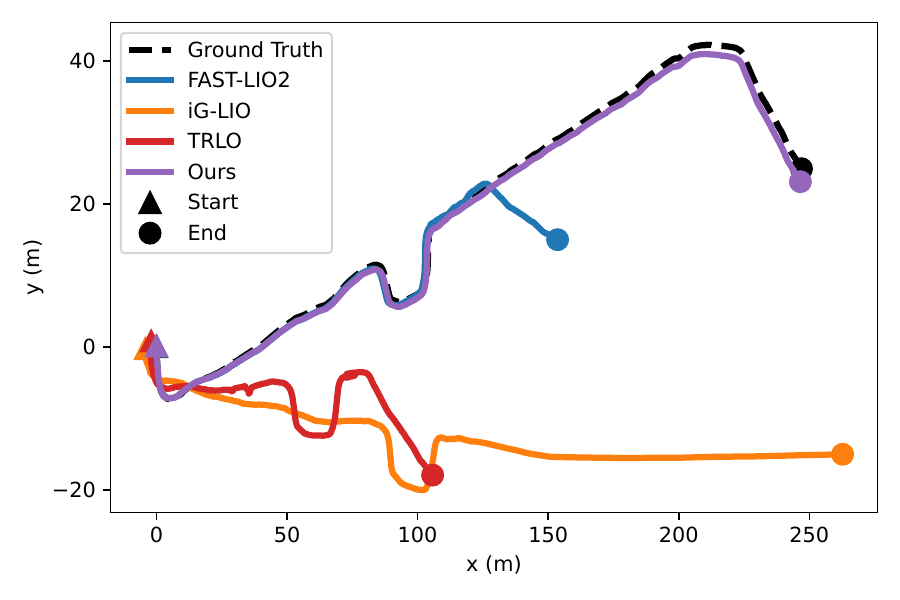}
    \includegraphics[width=0.32\textwidth]{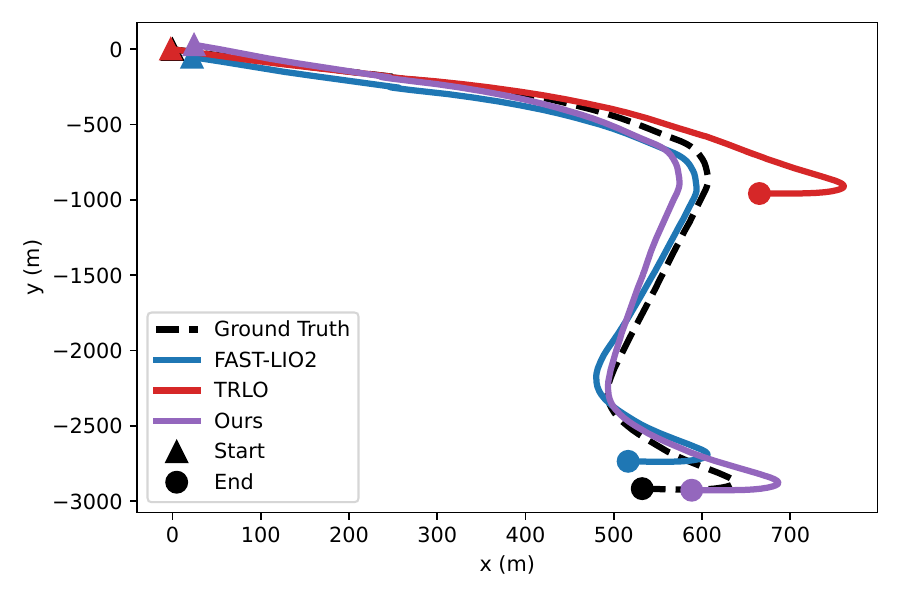}
    \includegraphics[width=0.32\textwidth]{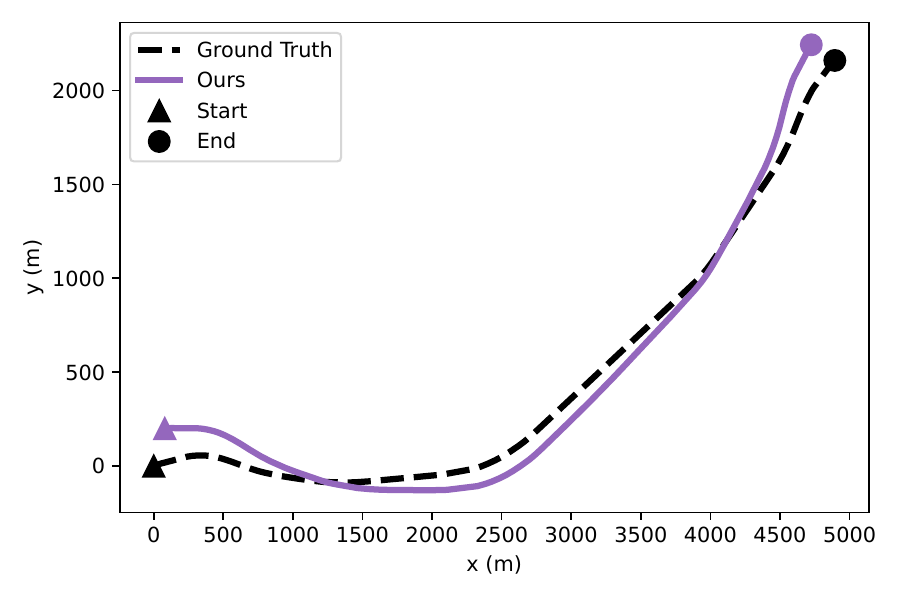}
    \vspace{-0.3cm} 
    \caption{Scatter plots of localization errors and corresponding trajectories in geometrically degenerate dynamic scenes. From left to right: results for sequences D\_3 G\_1, and G\_2. The first row shows RMSE scatter plots from five experiments, with methods having an average RMSE greater than 1000 marked by dashed boxes with diagonal lines. The second row shows ground truth and algorithm trajectories, excluding those with RMSE over 800.}
  \label{fig:GD_exp}
    \vspace{-0.3cm} 
\end{figure*}
\begin{table}[t]
\caption{Comparison of Absolute Trajectory Error (RMSE, in meters) in  Geometrically Rich Dynamic Scenes}
\centering{}%
\scalebox{0.85}{
\begin{threeparttable}
\begin{tabular}{cccccccc}
\toprule 
Method & D\_1 & D\_2 & E\_1 & E\_2 & E\_3 & U\_1 & U\_2\tabularnewline
\midrule
FAST-LIO2 & \colorbox{bshade!30}{2.15} & \XSolidBrush{} & \colorbox{bshade!30}{1.97} & 1.60 & \colorbox{bshade!30}{1.16} & 4.11 & 1.13\tabularnewline
IG-LIO & \colorbox{bshade!110}{1.98} & \XSolidBrush{} & \colorbox{bshade!110}{1.61} & \XSolidBrush{} & 1.18 & \colorbox{bshade!30}{3.91} & \colorbox{bshade!30}{0.92}\tabularnewline
Dynamic-LIO & - & - & 4.80 & \colorbox{bshade!110}{1.06} & 1.27 & 4.18 & 1.91\tabularnewline
TRLO & 3.37 & \colorbox{bshade!110}{0.54} & 8.82 & \colorbox{bshade!30}{1.23} & 1.21 & 11.96 & 0.94\tabularnewline
Ours & 3.65 & \colorbox{bshade!30}{0.64} & 1.98 & 1.47 & \colorbox{bshade!110}{0.89} & \colorbox{bshade!110}{3.43} & \colorbox{bshade!110}{0.91}\tabularnewline
\bottomrule
\end{tabular}
\begin{tablenotes}  
\item "-" indicates that the data for this sequence is not suitable for the algorithm to run. "\XSolidBrush{}" denotes that the algorithm diverges in this sequence, unable to obtain a meaningful trajectory. \parbox{1em}{\centering\colorbox{bshade!110}{}} and \parbox{1em}{\centering\colorbox{bshade!30}{}} represent the best and second-best results among all methods, respectively.
\end{tablenotes}
\end{threeparttable}
}

\label{table:gr_exp}
\vspace{-0.4cm} 
\end{table}

\subsubsection{Baselines and Metrics}
We benchmarked our method against four leading approaches: FAST-LIO2 \cite{xu2022fast} and iG-LIO \cite{chen2024ig}, which represent state-of-the-art LiDAR-inertial odometry for static environments, and Dynamic-LIO \cite{yuan2024lidar} and TRLO \cite{jia2024trlo}, which address dynamic environments through different strategies. Dynamic-LIO uses geometric features to detect moving objects, while TRLO employs learning-based detection with multi-object tracking before performing localization on the filtered point cloud.
For evaluation, we calculated the Root Mean Square Error (RMSE) between estimated and ground-truth trajectories after alignment using the evo package \cite{grupp2017evo}. To account for algorithmic randomness, we report the mean RMSE from five independent runs per sequence.

\subsubsection{Dataset}
We evaluated our method using a diverse dataset comprising both public benchmarks and custom recordings, as detailed in Table~\ref{Tab:dataset}. The dataset includes sequences from ECMD \cite{chen2023ecmd} (E\_x), UrbanNav \cite{hsu2021urbannav} (U\_x), GEODE \cite{chen2024heterogeneous} (G\_x), and our custom recordings (D\_x) with ground truth from RTK-GPS. The scenes of our custom dataset are illustrated in Fig. \ref{fig:dynamic_dataset_img}. These sequences were captured using various LiDAR sensors, including HDL-32E, VLP-16, and Livox Mid-360, representing both conventional spinning and solid-state scanning mechanisms.
To facilitate systematic evaluation, we categorized the sequences into three types based on environmental complexity. Geometrically-Rich (GR) environments contain abundant structural features with moderate dynamic elements, where conventional methods typically perform adequately. Geometrically-Degenerate (GD) environments present fewer distinct features and occasionally underconstrained scenarios (such as featureless tunnels), while still containing moving objects. Dynamic-Object-Dominated (DOD) environments primarily consist of moving objects, presenting the greatest challenge for state estimation.

\subsubsection{Geometrically Rich Benchmark}
Table \ref{table:gr_exp} shows localization performance across methods in GR environments. Most methods achieved comparable accuracy, with no clear advantage for dynamic-specific algorithms. Only in sequences D\_1 and D\_2, where moving objects significantly obstructed the sensor's view, did performance differences emerge.

These results highlight an important finding: in feature-rich environments, the structural elements provide sufficient constraints for accurate localization despite the presence of moving objects. Modern optimization techniques in LiDAR-inertial systems effectively treat points from dynamic objects as outliers when static features predominate. This explains why methods like iG-LIO perform well despite their static-world assumption. Consequently, evaluations limited to feature-rich environments may underestimate the benefits of dynamic object handling capabilities. A comprehensive assessment requires testing in more challenging conditions, particularly geometrically degenerate environments where dynamic objects have greater impact on localization performance, as demonstrated in the following section.

\subsubsection{Geometrically Degenerate Benchmark}
Fig. \ref{fig:GD_exp} (top row) presents results from GD environments. In these scenarios, moving objects significantly compromise methods that assume static scenes, as false point associations easily overwhelm the limited geometric constraints from surrounding structures. As shown in Fig.~\ref{fig:GD_exp} (bottom row, D\_3), FAST-LIO2's estimates can become immobilized while iG-LIO's trajectory gets distorted by dynamic objects.

While Dynamic-LIO and TRLO address dynamic environments, they separate object detection from point cloud registration. This creates an interdependence where accurate state estimation requires precise dynamic point segmentation and vice versa. Dynamic-LIO relies on ground segmentation and static map comparisons, but pose errors can corrupt this map, creating a cascading failure. TRLO uses frame-by-frame deep learning detection independent of registration, potentially causing inconsistent classifications across frames. Both approaches can experience trajectory drift when detection errors affect mapping.

Our approach resolves this interdependence by integrating dynamic point detection directly into registration. We filter moving objects during registration by modeling point velocity through spatiotemporal normal vectors calculated across all points within a defined time window. This avoids reliance on static map references and prevents error propagation. Our algorithm successfully estimates sensor trajectory across all sequences, consistently outperforming other methods. This demonstrates the clear advantage of our integrated approach to dynamic object handling.

\begin{table}
\caption{Comparison of Absolute Trajectory Error (RMSE, in meters) in Dynamic-Object Dominated Scenes}

\centering{}%
\resizebox{0.48\textwidth}{!}{
\begin{threeparttable}
\begin{tabular}{cccc}
\toprule 
\makebox[0.1\textwidth][c]{name} & \makebox[0.1\textwidth][c]{D\_4} & \makebox[0.1\textwidth][c]{D\_5}
                                     & \makebox[0.1\textwidth][c]{D\_6} \tabularnewline
\midrule
FAST-LIO & 12.40 & 28.33 & 16.99\tabularnewline
IG-LIO & \colorbox{bshade!110}{0.74} & \XSolidBrush{} & \XSolidBrush{}\tabularnewline
Dynamic-LIO & - & - & -\tabularnewline
TRLO & 0.80 & \colorbox{bshade!30}{1.91} & \colorbox{bshade!110}{2.82}\tabularnewline
Ours & \colorbox{bshade!30}{0.79} & \colorbox{bshade!110}{1.83} & \colorbox{bshade!30}{3.04}\tabularnewline
\bottomrule
\end{tabular}
\begin{tablenotes}  
\item "-" indicates that the data for this sequence is not suitable for the algorithm to run. "\XSolidBrush{}" denotes that the algorithm diverges in this sequence, unable to obtain a meaningful trajectory. \parbox{1em}{\centering\colorbox{bshade!110}{}} and \parbox{1em}{\centering\colorbox{bshade!30}{}} represent the best and second-best results among all methods, respectively..
\end{tablenotes}
\end{threeparttable}
}
\label{table:dod_exp}
\vspace{-0.4cm} 
\end{table}

\subsubsection{Dynamic-Object Dominated Benchmark}
Table \ref{table:dod_exp} shows odometry results for DOD sequences. Despite containing structural features, these environments present significant challenges as numerous moving elements frequently obscure static landmarks.

Methods lacking dynamic object handling capabilities (FAST-LIO and iG-LIO) demonstrated poorer trajectory accuracy throughout most sequences. While these approaches successfully maintained localization in the simpler D\_4 scenario, they failed in the more challenging D\_5 and D\_6 environments. In contrast, our approach and TRLO, which explicitly account for moving elements, maintained reliable localization across all test cases. These results highlight the critical importance of robust dynamic object handling for accurate state estimation in highly dynamic settings.

\subsection{Static Map Building}
To evaluate our static map building capability in dynamic environments, we assessed our dynamic point filtering performance against several baseline methods. We compared with offline approaches (BeautyMap \cite{Jia2024BeautyMapBA}, DUFOMap \cite{duberg2024dufomap}), a delayed incremental method \cite{legentil2024realtime}, and online techniques (Dynablox \cite{schmid2023dynablox}, OTD \cite{Wu2024ObservationTD}). For fair comparison, all methods used ground-truth poses. 
We conducted quantitative evaluation using the pointwise protocol from \cite{zhang2023benchmark} on nine sequences from the Helimos dataset \cite{lim2024helimos}, which features highly dynamic scenes captured by four different LiDAR sensors (Avia, AEVA, VLP16, and Ouster128). Our evaluation metrics include Dynamic Accuracy (DA) and Static Accuracy (SA), measuring the recall of dynamic and static points respectively, and their harmonic mean (HA = $\frac{2 \times \text{SA} \times \text{DA}}{\text{SA} + \text{DA}}$) as a balanced overall measure.

\begin{table}
\begin{centering}
\caption{QUANTITATIVE COMPARISON OF DYNAMIC POINTS REMOVAL IN POINT CLOUD MAPS}
\label{tab:static_map_building}
\par\end{centering}
\centering{}%
\scalebox{0.79}{
\begin{threeparttable}
\begin{tabular}{c@{ }c@{ }|c@{  }c@{  }|c|ccc}
\toprule 
\multicolumn{2}{c}{Sensor \textbackslash{} Method} & Beautymap & DUFOMap & MCDOD & Dynablox & OTD & Ours\tabularnewline
\midrule 
\multirow{3}{*}{Aeva} & SA {[}\%{]} \textuparrow{} & 69.34 & \colorbox{bshade!30}{91.93} & 61.60 & \colorbox{bshade!110}{\textbf{97.64}} & 80.74 & \underline{91.50} \tabularnewline
 & DA {[}\%{]} \textuparrow{} & \colorbox{bshade!30}{79.98} & 79.33 & \colorbox{bshade!110}{94.71} & \underline{64.78} & 43.22 & \textbf{77.77}\tabularnewline
 & HA {[}\%{]} \textuparrow{} & 74.12 & \colorbox{bshade!110}{84.57} & 74.21 & \underline{75.81} & 54.26 & \colorbox{bshade!30}{\textbf{82.50}}\tabularnewline
\midrule 
\multirow{3}{*}{Avia} & SA {[}\%{]} \textuparrow{} & 79.77 & \colorbox{bshade!110}{96.30} & 74.42 & \colorbox{bshade!30}{\textbf{95.21}} & \underline{93.80} & 93.66\tabularnewline
 & DA {[}\%{]} \textuparrow{} & \colorbox{bshade!110}{85.30} & 72.33 & \colorbox{bshade!30}{83.12} & 59.54 & \textbf{76.66} & \underline{67.64}\tabularnewline
 & HA {[}\%{]} \textuparrow{} & \colorbox{bshade!30}{81.51} & 80.43 & 77.76 & 69.42 & \colorbox{bshade!110}{\textbf{83.20}} & \underline{75.87}\tabularnewline
\midrule 
\multirow{3}{*}{Ouster} & SA {[}\%{]} \textuparrow{} & 92.24 & 90.06 & 77.70 & \colorbox{bshade!110}{\textbf{97.02}} & 92.09 & \colorbox{bshade!30}{\underline{95.46}}\tabularnewline
 & DA {[}\%{]} \textuparrow{} & 86.72 & \colorbox{bshade!30}{90.23} & \colorbox{bshade!110}{91.34} & \underline{71.68} & 63.64 & \textbf{72.07}\tabularnewline
 & HA {[}\%{]} \textuparrow{} & \colorbox{bshade!30}{89.27} & \colorbox{bshade!110}{89.99} & 83.75 & \textbf{82.09} & 74.70 & \underline{81.80}\tabularnewline
\midrule 
\multirow{3}{*}{Velodyne} & SA {[}\%{]} $\uparrow$ & 76.62 & \colorbox{bshade!30}{85.18} & 46.18 & \colorbox{bshade!110}{\textbf{97.62}} & \underline{84.56} & 74.15\tabularnewline 
 & DA {[}\%{]} \textuparrow{} & \colorbox{bshade!30}{84.73} & 13.12 & \colorbox{bshade!110}{85.40} & 31.87 & \underline{45.73} & \textbf{48.93}\tabularnewline 
 & HA {[}\%{]} \textuparrow{} & \colorbox{bshade!110}{80.10} & 21.58 & \colorbox{bshade!30}{59.36} & 47.33 & \textbf{58.41} & \underline{57.79}\tabularnewline
\bottomrule
\end{tabular}
\begin{tablenotes}  
\item \parbox{1em}{\centering\colorbox{bshade!110}{}} and \parbox{1em}{\centering\colorbox{bshade!30}{}} represent the best and second-best results among all methods, respectively. Bold and underlined text indicate the best and second-best results among online methods, respectively.
\end{tablenotes}
\end{threeparttable}
}
\vspace{-0.4cm} 
\end{table}

\begin{figure}[t]
  \centering
  \includegraphics[width=0.48\textwidth]{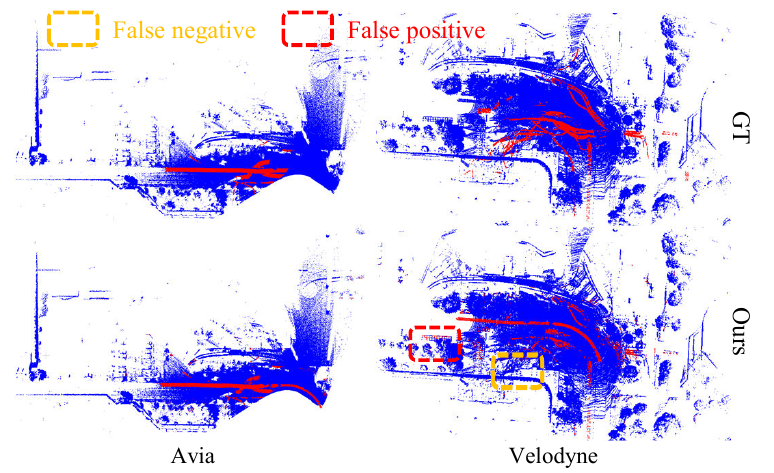}
  \caption{Static Map Construction Comparison Across LiDAR Sensors (Red: Dynamic, Blue: Static)}
  \label{fig:qualitative_vis_img}
  \vspace{-0.4cm} 
\end{figure}

Table \ref{tab:static_map_building} compares static map building performance across all test sequences. Our method achieves optimal online results with the Aeva sensor, nearly matching the best offline approaches. For other LiDAR types, we demonstrate the second-best online performance, showing consistent mapping capabilities across different sensors.
Fig. \ref{fig:qualitative_vis_img} demonstrates how sensor characteristics influence performance. The effectiveness of our approach depends on point cloud density and distribution, which varies between sensor models. Sensors with sparse point distributions challenge the assumption of local uniformity needed for accurate normal vector estimation, increasing false positive rates. This explains the performance differences observed across sensor types in Table \ref{tab:static_map_building}.
While our spatial consistency check effectively reduces false positives and improves static accuracy compared to MCDOD \cite{legentil2024realtime}, sparse dynamic points that fail to form clear clusters may remain undetected. This accounts for the lower dynamic accuracy with certain sensor configurations compared to MCDOD, highlighting a trade-off between static and dynamic point classification.


\subsection{Ablation Experiments}

To assess the contribution of each component in our unified framework for dynamic point handling and registration, we conducted an ablation study on three representative sequences. We compared the full system with two simplified variants:

\begin{itemize}
    \item \textbf{Baseline (No Dynamic Detection)}: A simplified version that processes all points as static, equivalent to conventional LIO methods without dynamic handling.
    \item \textbf{Sequential Processing}: A version that first detects dynamic points using our spatiotemporal normal vector method based on initial pose estimates, then performs registration using only the filtered point cloud.
\end{itemize}

\begin{table}[t]
\caption{Ablation Study: RMSE Comparison of Different Variants of Our Method on Representative Sequences}
\label{table:ablation}
\begin{centering}
\begin{tabular}{cccc}
\toprule 
 & Ours w/o Dyn. & Ours w/ Sep. Det. & Ours w/ Full\tabularnewline
\midrule 
GR(D\_2) & 0.79 & \textbf{0.59} & 0.64\tabularnewline
GD(P\_3) & 35.84 & 0.60 & \textbf{0.46}\tabularnewline
DOD(D\_4) & 6.14 & 1.01 & \textbf{0.79}\tabularnewline
\bottomrule
\end{tabular}
\par\end{centering}
\vspace{-0.4cm} 
\end{table}

Table~\ref{table:ablation} shows that our full approach achieves the lowest RMSE in challenging GD and DOD scenarios. In the geometrically-rich sequence (D\_2), the sequential detection variant performs slightly better, likely because abundant static features provide reliable initial pose estimates for detection. However, the substantial performance degradation when dynamic point detection is disabled demonstrates the critical importance of accounting for moving objects. Our unified approach delivers superior accuracy in geometrically degenerate and dynamic-dominated scenarios where initial pose estimates may be unreliable. These results confirm the effectiveness of integrating dynamic point handling directly within the registration pipeline, particularly for challenging environments.

\subsection{Processing Time Analysis}

We analyzed the computational efficiency of our approach using the \textit{Promenade03} sequence. Our method achieved an average processing time of $49.69\,\text{ms}$ per LiDAR scan, well below the typical $100\,\text{ms}$ scan interval, demonstrating real-time performance. The core dynamic-aware state estimation component, including all normal vector computations, required $37.38\,\text{ms}$, while our spatial consistency check for dynamic point filtering added only $6.08\,\text{ms}$ of overhead. These results confirm that our approach maintains computational efficiency while handling dynamic environments effectively.
\section{CONCLUSION}
This paper presents a dynamic-aware Lidar-Inertial Odometry framework that effectively addresses localization challenges in dynamic environments. By integrating spatio-temporal normal analysis into point cloud registration, our approach explicitly handles dynamic objects while resolving the circular dependency between state estimation and dynamic object detection that has constrained previous methods.
Experimental results on both public datasets and our newly introduced dynamic environment dataset demonstrate superior performance across diverse scenarios, particularly in geometrically degraded scenes where conventional LIO systems struggle. Our spatial consistency verification method enhances map quality while maintaining computational efficiency, enabling real-time operation on standard hardware.



\ifCLASSOPTIONcaptionsoff
  \newpage
\fi



%
\bibliographystyle{IEEEtran}
\bibliography{main}

%








\end{document}